\documentclass[10pt,twocolumn,letterpaper]{article}

\usepackage{iccv}
\usepackage{times}
\usepackage{epsfig}
\usepackage{graphicx}
\usepackage{amsmath}
\usepackage{amssymb}

\usepackage{graphicx}
\usepackage{footmisc}
\usepackage{booktabs}       
\usepackage{amsfonts}       
\usepackage{nicefrac}       
\usepackage{microtype}      
\usepackage{color}
\usepackage{bm}
\usepackage{amsmath}
\usepackage{wrapfig}
\usepackage{graphicx}
\usepackage{setspace}
\usepackage[super]{nth}
\usepackage{siunitx}

\usepackage{caption}

\usepackage{blindtext}
\usepackage{marvosym}
\usepackage{times}
\usepackage{epsfig}
\usepackage{graphicx}
\usepackage{amsmath}
\usepackage{amssymb}
\usepackage{colortbl}
\definecolor{mygray}{gray}{.88}
\usepackage{latexsym}
\usepackage{CJKutf8}

\usepackage[pagebackref=true,breaklinks=true,colorlinks,bookmarks=false]{hyperref}
\usepackage{hyperref}
\usepackage{bm}

\usepackage{caption}
\usepackage[caption=false]{subfig}
\usepackage{multirow}
\usepackage{multicol}
\usepackage{adjustbox}

\usepackage{pifont}
\usepackage{xspace}
\newcommand{\cmark}{\ding{51}\xspace}%
\newcommand{\xmarkg}{\textcolor{lightgray}{\ding{55}}\xspace}%

\usepackage{bm}

\usepackage[pagebackref=true,breaklinks=true,colorlinks,bookmarks=false]{hyperref}

\iccvfinalcopy 

\def\iccvPaperID{1578} 

\ificcvfinal\fi

\begin{document}

\title{Deep Geometrized Cartoon Line Inbetweening}

\author{Li Siyao$^1$ \ \ \ \ Tianpei Gu$^{2*}$ \ \ \ \ Weiye Xiao$^3$ \ \ \ \ Henghui Ding$^1$ \ \ \ \ Ziwei Liu$^1$ \ \ \ \ Chen Change Loy\textsuperscript{1~\Letter}\\
$^1$S-Lab, Nanyang Technological University \ \ \ \ $^2$Lexica \ \ \ \ $^3$Southeast University\\
{\tt\footnotesize \{siyao002, henghui.ding, ziwei.liu, ccloy\}@ntu.edu.sg, gutianpei@ucla.edu, 230189776@seu.edu.cn}
}

\twocolumn[{%
\renewcommand\twocolumn[1][]{#1}%
\maketitle
\begin{center}
    \centering
    \small{
     \vspace{-4pt}
    \includegraphics[width=0.92\linewidth, trim=0pt 0pt 0pt 30pt, clip]{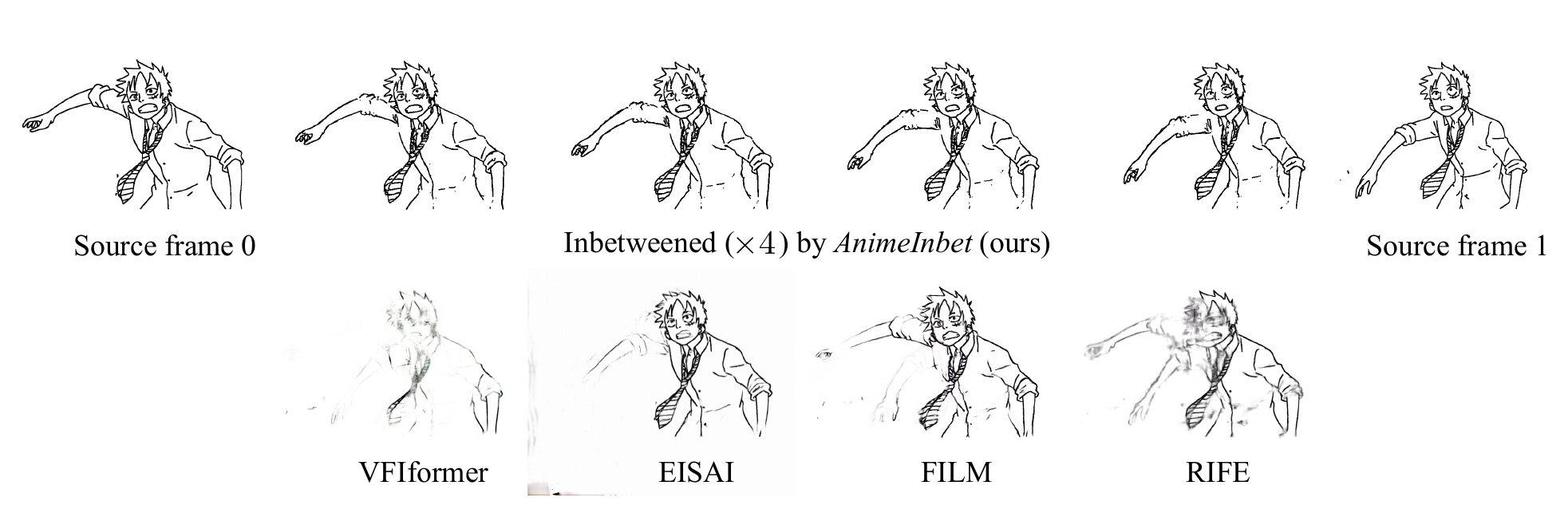}
        
    \vspace{-2pt}
    \captionof{figure}{\small{\textbf{Inbetweening on two source cartoon line drawings of \textit{Monkey D. Luffy}  extracted from  \textit{ONE PIECE}}. We compare our proposed \textit{AnimeInbet} with state-of-the-art frame interpolation methods VFIformer \cite{lu2022video}, EISAI \cite{chen2022improving}, FILM \cite{reda2022film} and RIFE \cite{huang2022rife}.}} 

    \vspace{4pt}
    \label{fig:teaser}
    }
\end{center}%
}]

\maketitle

\newcommand\blfootnote[1]{%
\begingroup
\renewcommand\thefootnote{}\footnote{#1}%
\addtocounter{footnote}{-1}%
\endgroup
}

\blfootnote{\textsuperscript{\Letter}~Corresponding author. \  $^*$Work completed at UCLA.}
\ificcvfinal\thispagestyle{empty}\fi


\begin{abstract}
\vspace{-4pt}
    We aim to address a significant but understudied problem in the anime industry, namely the inbetweening of cartoon line drawings. Inbetweening involves generating intermediate frames between two black-and-white line drawings and is a time-consuming and expensive process that can benefit from automation. However, existing frame interpolation methods that rely on matching and warping whole raster images are unsuitable for line inbetweening and often produce blurring artifacts that damage the intricate line structures.
    To preserve the precision and detail of the line drawings, we propose a new approach,  AnimeInbet, which geometrizes raster line drawings into graphs of endpoints and reframes the inbetweening task as a graph fusion problem with vertex repositioning. 
    Our method can effectively capture the sparsity and unique structure of line drawings while preserving the details during inbetweening. {This is made possible via our novel modules, \ie, vertex geometric embedding, a vertex correspondence Transformer, an effective mechanism for vertex repositioning and a visibility predictor.}
    %
    %
    To train our method, we introduce MixamoLine240, a new dataset of line drawings with ground truth vectorization and matching labels. Our experiments demonstrate that AnimeInbet synthesizes high-quality, clean, and complete intermediate line drawings, outperforming existing methods quantitatively and qualitatively, especially in cases with large motions.
    Data and code are available at \url{https://github.com/lisiyao21/AnimeInbet}.
    \vspace{-4pt}

\end{abstract}


\vspace{-6pt}
\section{Introduction}
\label{sec:intro}

Cartoon animation has undergone significant transformations since its inception in the early 1900s, when consecutive frames were manually drawn on paper. 
%
Although automated techniques now exist to assist with some specific procedures during animation production, such as colorization \cite{qu2006manga,sykora2004unsupervised,kopf2012digital,zhang2018two,casey2021animation} and special effects \cite{zhang2021shadowing}, the core element -- the line drawings of characters -- still needs hand-drawing each frame individually, making 2D animation a labor-intensive industry.
Developing an automated algorithm that can produce intermediate line drawings from two input key frames, commonly referred to as ``inbetweening'', has the potential to significantly improve productivity. 

\begin{figure}[t]

    \setlength{\tabcolsep}{1.5pt}
    \centering
    \includegraphics[width=0.88\linewidth]{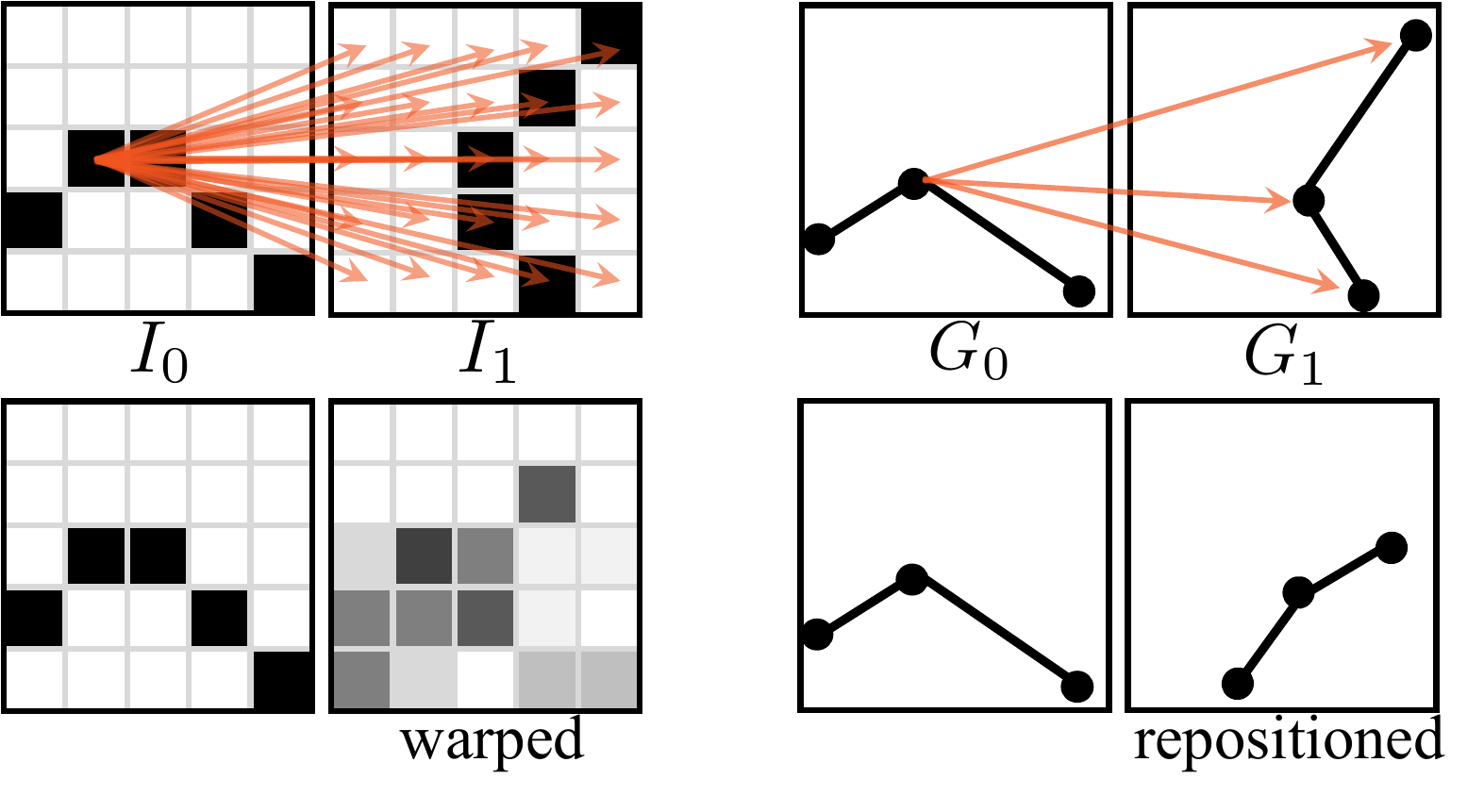}

    \vspace{-2pt}
    \caption{\small{\textbf{Raster  \textit{vs} geometrized  inbetweening.} Top: search space of a pixel (left) \textit{vs} a vertex (right) in matching.
    Bottom: pixel warping/sampling (left) \textit{vs} vertex repositioning (right).
    }}
    \label{fig:raster_vector}

\end{figure}

Line inbetweening is not a trivial subset of general frame interpolation, as the structure of line drawings is extremely sparse. Unlike full-textured images, line drawings contain only around 3\% black pixels, with the rest of the image being white background. As illustrated in Figure \ref{fig:raster_vector}, this poses two significant challenges for existing raster-image-based frame interpolation methods.
\textbf{1)} The lack of texture in line drawings makes it challenging to compute pixel-wise correspondence accurately in frame interpolation. One pixel can have many similar matching candidates, leading to inaccurate motion prediction.
\textbf{2)} The warping and blending used in frame interpolation can blur the salient boundaries between the line and the background, leading to a significant loss of detail. 

To address the challenges posed by line inbetweening, we propose a novel deep learning framework called \emph{AnimeInbet}, which inbetweens line drawings in a geometrized format instead of raster images. Specifically, the source images are transformed into vector graphs, and the goal is to synthesize an intermediate graph. This reformulation can overcome the challenges discussed earlier in this paper.
As illustrated in Figure \ref{fig:raster_vector}, the matching process in the geometric domain is conducted on concentrated geometric endpoint vertices, rather than all pixels, reducing potential ambiguity and leading to more accurate correspondence. Moreover, the repositioning does not change the topology of the line drawings, enabling preservation of the intricate and meticulous line structures.
Compared to existing methods, our proposed \emph{AnimeInbet} framework can generate clean and complete intermediate line drawings, as demonstrated in Figure \ref{fig:teaser}. 

The core idea of our proposed \emph{AnimeInbet} framework is to find matching vertices between two input line drawing graphs and then reposition them to create a new intermediate graph. To achieve this, we first design a vertex encoding strategy that embeds the geometric features of the endpoints of sparse line drawings, making them distinguishable from one another. We then apply a vertex correspondence Transformer to match the endpoints between the two input line drawings.
Next, we propagate the shift vectors of the matched vertices to unmatched ones based on the similarities of their aggregated features to realize repositioning for all endpoints. Finally, we predict a visibility mask to erase the vertices and edges occluded in the inbetweened frame, ensuring a clean and complete intermediate frame.

To facilitate supervised training on vertex correspondence, we introduce \emph{MixamoLine240}, the first line art dataset with ground truth geometrization and vertex matching labels. The 2D line drawings in our dataset are selectively rendered from specific edges of a 3D model, with the endpoints indexed from the corresponding 3D vertices. By using 3D vertices as reference points, we ensure that the vertex matching labels in our dataset are accurate and consistent at the vertex level.

%
%

In a conclusion, our work contributes a new and challenging task of line inbetweening, which could facilitate one of the most labor-intensive art production processes. We also propose a new method that outperforms existing solutions, and introduce a new dataset for comprehensive training.



\section{Related Work}
\label{sec:related}

\begin{figure}[t]

    \setlength{\tabcolsep}{1.5pt}
    \centering
    \includegraphics[width=0.996\linewidth]{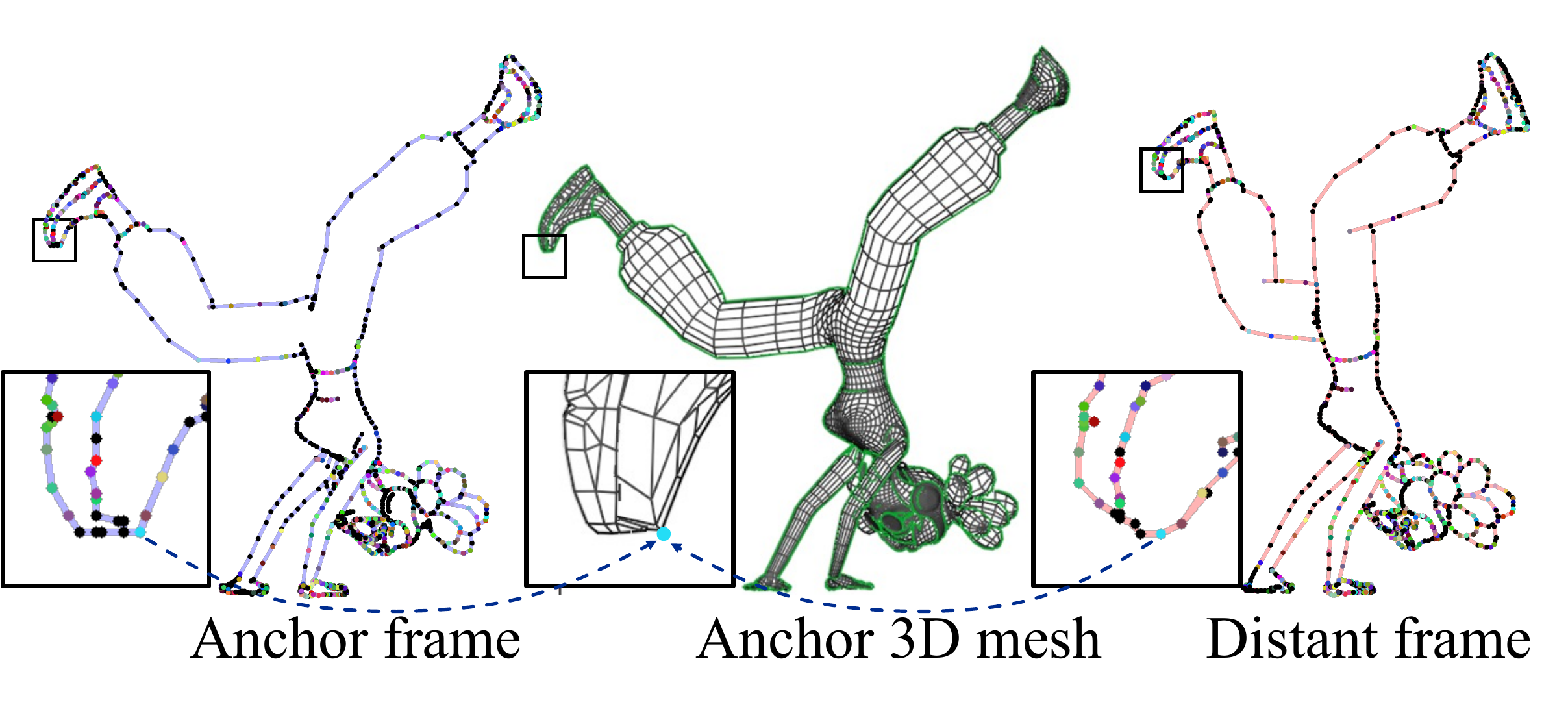}

    \vspace{-3pt}
    \caption{\small{\textbf{Geometrized line art in \textit{MixamoLine240}.} 2D endpoints and connected lines are projected from vertices and edges of orinal 3D mesh.} Endpoints indexed to unique 3D vertices are matched (marked in the same colors).} 
    \label{fig:index_to_3D}

\end{figure}

\noindent{\textbf{Frame Interpolation.}}
Frame interpolation is a widely studied task in recent years, involving synthesizing intermediate frames from existing ones. Many approaches have been proposed~\cite{liu2017video,niklaus2017video,niklaus2017video2,jiang2018super,niklaus2018context,xu2019quadratic,niklaus2020softmax,park2020bmbc,sim2021extreme,huang2022rife,reda2022film,chen2022improving,lu2022video,li2020deep}, such as those that use optical flows or deep networks to search for matching areas and warp them to proper intermediate locations.
Among the most recent algorithms, RIFE \cite{huang2022rife} directly predicts intermediate flows to warp the input frames and blends the warped frames into intermediate ones by a visible mask. VFIformer \cite{lu2022video} adopts the same idea to predict the intermediate flows but proposes a Transformer to synthesize the intermediate from both warped images and features.
Reda \etal \cite{reda2022film} design a scale-agnostic feature pyramid to  predict the intermediate flows and warp frames in a hierarchical manner to handle extreme large motions.
Siyao and Zhao \etal \cite{siyao2021avi} propose a frame interpolation pipeline specific for 2D cartoon in the wild, while Chen and Zwicker \cite{chen2022improving} improves the perceptual quality by embedding an optical-flow based line aggregator.
While these methods achieve impressive performance on raster natural or cartoon videos, their pixel-oriented nature are not suitable for inbetweening concise and sparse line arts, which can yield severe artifacts and are not feasible for real usage in anime creation.

\noindent{\textbf{Research on Anime.}}
There has been increasing research interest in techniques to facilitate 2D cartoon creation, including sketch simplification  \cite{simo2016learning,simo2018mastering}, vectorization \cite{zhang2009vectorizing,yao2016manga,virtualskethcer,liu2022end}, colorization \cite{qu2006manga,sykora2004unsupervised,kopf2012digital,zhang2018two,casey2021animation}, shading \cite{zhang2021shadowing}, head reenactment \cite{kim2021animeceleb} and line-art-based cartoon generation \cite{zhang2023adding}.
%
While these studies may improve specific aspects of animation creation, the core line arts still rely on manual frame-by-frame drawing. Some sporadic rule-based methods have been developed for stroke inbetweening under strict conditions, but these methods lack the flexibility required for wider applications \cite{yang2017context,carvalho2017dilight}. Our work is the first to propose a deep learning-based method for inbetweening geometrized line arts. Additionally, we introduce vertex-wise correspondence datasets on line arts. It is noteworthy that existing datasets are not sufficiently `clean' for our task since cartoon contour lines can cross the boundaries of motion, leading to incorrect corresponding labels at the vertex level \cite{shugrina2019creative,siyao2022animerun}.

\section{Mixamo Line Art Dataset}
\label{sec:data}

\begin{figure}[t]

    \setlength{\tabcolsep}{2pt}
\centering
\subfloat{    \includegraphics[width=0.996\linewidth]{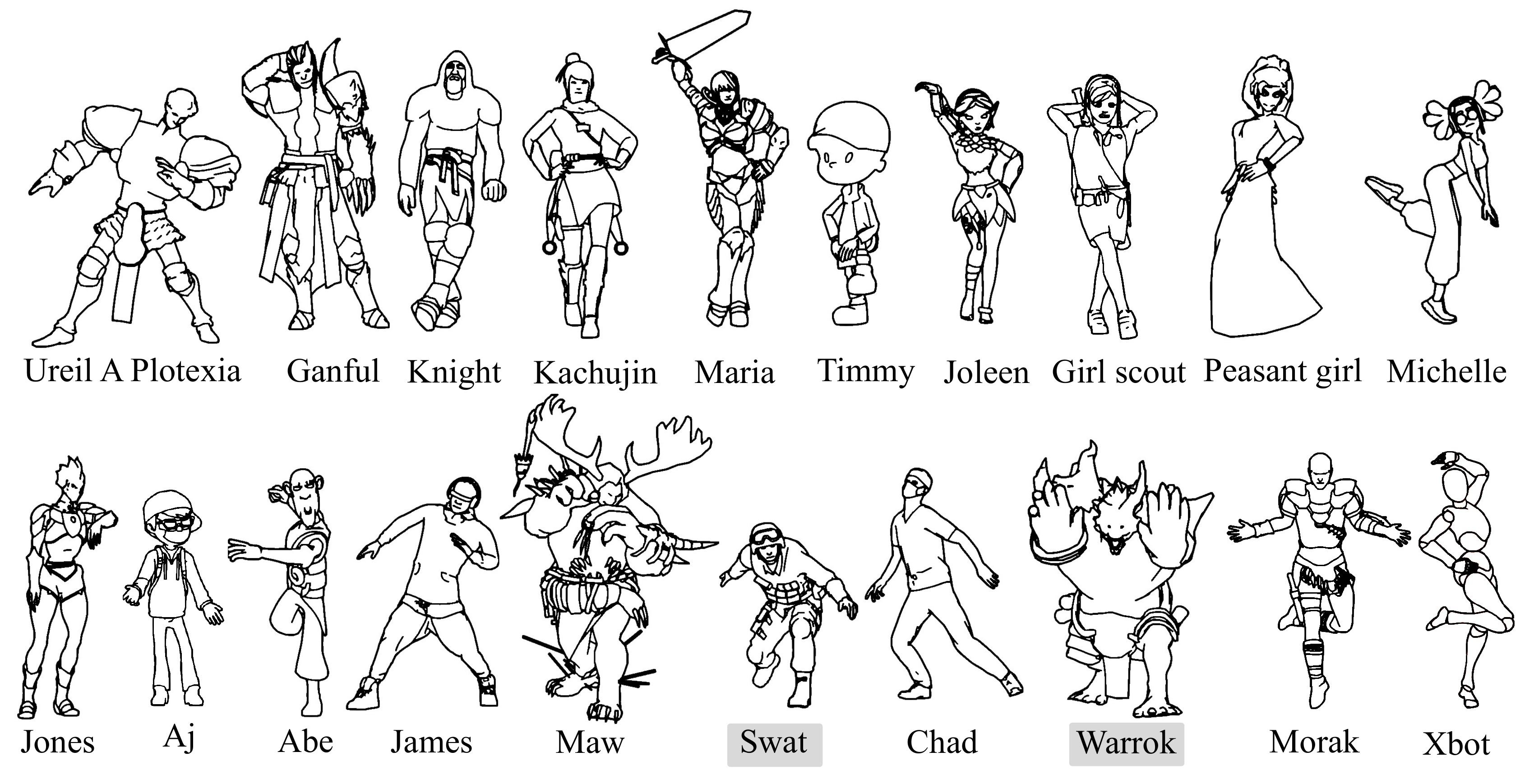}
}
\vspace{-2pt}
\subfloat{
  \scriptsize
{
  \begin{tabular}{l  c c c c c c c  c  c c }
    \toprule
    \textbf{Train} & break dance & capoeira & chap giratoria & fist fight & flying knee \\
    \textbf{actions} & climb & run & shove & magic attack & tripping\\

      \midrule
      
    \textbf{Test} & chip & evade & flair & \cellcolor{gray!30} sword slash &  \cellcolor{gray!30} hip hop \\
    \textbf{actions}& hurricane kick & soccer tackle & standing death & swim & stand up\\

    \bottomrule
  \end{tabular}
  }
}

    \vspace{-2pt}
    \caption{\small{\textbf{Data composition.} Training and test sets are separately composed by 10 characters $\times$ 10 actions. First \& second rows are training \& test characters, respectively. Shaded are for validation. }}
    
    \label{fig:character_action}

\end{figure}

To facilitate training and evaluation of geometrized line inbetweening, we develop a  large-scale dataset, named \textit{MixamoLine240}, which consists of 240 sequences of consecutive line drawing frames, with 100 sequences for training and 140 for validation and testing. 
To obtain this vast amount of cartoon line data, we utilize a ``Cel-shading'' technique, \ie, to use computer graphics software (Blender in this work) to render 3D resources into an anime-style appearance that mimics the hand-drawn artistry.
Unlike previous works \cite{shugrina2019creative,siyao2022animerun} that only provide raster images, \textit{MixamoLine240} also provides ground-truth geometrization labels for each frame, which include the coordinates of a group of vertices ($V$) and the connection topology ($T$). 
Additionally, we assign an index number ($R[i]$) to each 2D endpoint ($V[i]$) that refers to a unique vertex in the 3D mesh of the character, as illustrated in Figure \ref{fig:index_to_3D}, which can be further used to deduce the vertex-level correspondence.
Specifically, given two frames $I_0$ and $I_1$ in a sequence, the 3D reference IDs reveal the vertex correspondence  $ \{(i,j)\}$ for those vertices $i$ in $I_0$ and $j$ in $I_1$ having $R_0[i] = R_1[j]$, while the rest unmatched vertices are marked as occluded. 
This strategy allows us to produce correspondence pairs with arbitrary frame gaps to flexibly adjust the input frame rate during training.
Next, we discuss the construction and challenges inherent in the data. 



\noindent\textbf{Data Construction.}
In Blender, the mesh structure of a 3D character remains stable, \ie, the number of 3D vertex and the edge topology keep constant, when moving without additional subdivision modifier. We employ this property to achieve consistent line art rendering and accurate annotations for geometrization and vertex matching.
As shown in Figure \ref{fig:index_to_3D}, the original 3D mesh contains all the necessary line segments required to represent the character in line art. During rendering, the visible outline from the camera's perspective is selected based on the material boundary and the object's edge. This process ensures that every line segment in the resulting raster image corresponds to an edge in the original mesh. The 2D endpoints of each line segment are simply the relevant 3D vertices projected onto the camera plane, referenced by the unique and consistent index of the corresponding 3D vertex.
%
%
%
Meanwhile, since the 3D mesh naturally defines the vertex connections, the topology of the 2D lines can be transferred from the selective edges used for rendering.
To prevent any topological ambiguity that may be caused by overlapped vertices in 3D space, we merge the endpoints that are within a Euclidean distance of $0.1$ in the projected 2D space. This enables us to obtain both the raster line drawings and the accurate labels of each frame.

\begin{table}
    \setlength{\tabcolsep}{3pt}
    \renewcommand\arraystretch{1}
\caption{\small\textbf{Difficulty statistics with various frame gaps.}
}
\vspace{-4pt}
  
  \centering
  \footnotesize
{
  \begin{tabular}{c l  c c c c    }
    \toprule
    
 & {Frame gap}$\to$ &  0 (60 fps) & 1 (30 fps) & 5 (10 fps)  & 9 (6 fps) \\
      \midrule
    \multirow{3}[0]*{\rotatebox{90}{Train}} & Occlusion rate (\%) & 14.8 & 21.5 & \ \ 37.8 & \ \ 46.6  \\
    & Avg. vtx shift & \ \ 8.6 & 16.4 & \ \ 42.6 & \ \ 62.8 \\
    & Avg. max vtx shift & 26.0 & 48.9 & 129.7 & 192.3\\
    \midrule
    \multirow{3}[0]*{\rotatebox{90}{Test}} & Occlusion rate (\%) & 18.4 & 26.5 & \ \ 44.2 & \ \ 53.5  \\
    & Avg. vtx shift  & \ \ 7.8 & 14.9 & \ \ 38.9 & \ \ 57.0 \\
    & Avg. max vtx shift  & 23.8 & 45.0 & 119.3 & 173.5\\
    \bottomrule
  \end{tabular}
  }
  \label{table:statistics}
\end{table}

\begin{figure*}[t]

    \setlength{\tabcolsep}{1.5pt}
    \centering
    \includegraphics[width=0.996\linewidth]{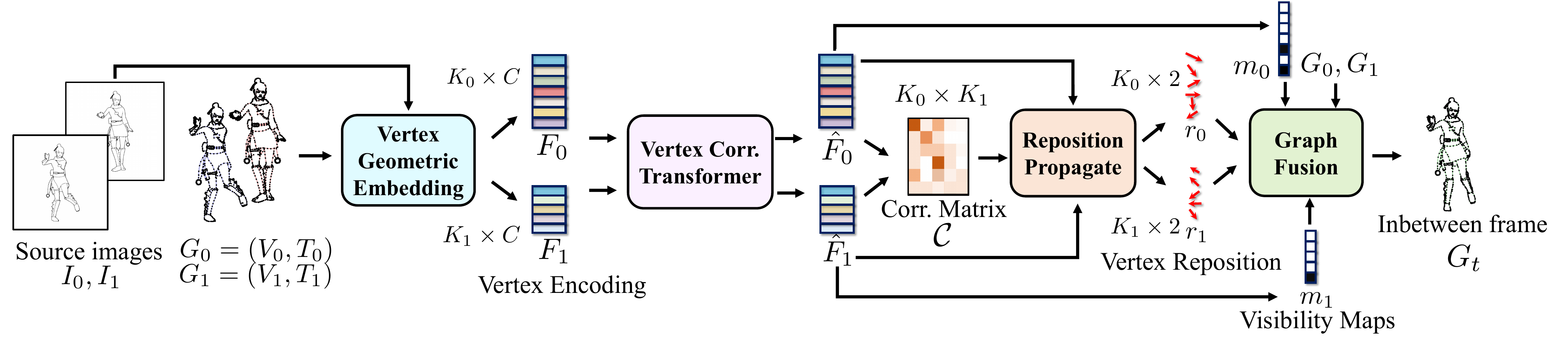}

    \vspace{-4pt}
    \caption{\small{\textbf{Pipeline of proposed \emph{AnimeInbet}.} 
    Our framework is composed of four main parts: the vertex geometric embedding, the vertex correspondence Transformer, repositioning propagation and graph fusion.
    Given a pair of line images $I_0$ and $I_1$ and their vector graphs $G_0$ and $G_1$, our method generates the intermediate frame $G_t$ in geometrized format.
    }} 
    \label{fig:pipeline}

\end{figure*}

To create a diverse dataset, we used the open-source 3D material library Mixamo \cite{mixamo} and selected 20 characters and 20 actions, as shown in Figure \ref{fig:character_action}. Each action has an average of 191 frames. We combined 10 characters and 10 actions to render 100 sequences, with a total of 19,930 frames as the training set. We then used the remaining 10 characters and 10 actions to render an 18,230-frame test set, ensuring that the training and testing partitions are exclusive.
We also created a 44-sequence validation set, consisting of 20 unseen characters, 20 unseen actions, and 4 with both unseen character and action. To create this set, we combined the test characters ``Swat''and ``Warrok'' and actions  ``sword slash'' and ``hip hop'' with the training characters and actions. The validation set contains 11,102 frames and was also rendered at 1080p resolution with a frame rate of 60 fps. To ensure consistency across all frames, we cropped and resized each frame to a unified $720\times720$ character-centered image.

\noindent\textbf{Challenges.}
%
%
Table \ref{table:statistics} summarizes the statistics that reflect the difficulty of the line inbetweening task under various input frame rates. With an increase in frame gaps, the inbetweening task becomes more challenging with larger motion magnitudes and higher occlusion percentages. For instance, when the frame gap is 9, the input frame rate becomes 6 fps, and the average vertex shift is 62.8 pixels. The mean value of the maximum vertex shift in a frame (``Avg. max vtx shift'') reaches 192.3 pixels, which is  27\% of the image width. Additionally, nearly half of the vertices are unmatched in such cases, making line inbetweening a tough problem. Furthermore, the image composition of the test set is more complex than that of the training set. A training frame has an average of 1,256 vertices and 1,753 edges, while a test frame has an average of 1,512 vertices and 2,099 edges since the test set has more complex characters such as ``Maw''.



\section{Our Approach}
\label{sec:method}

An overview of the proposed line inbetweening framework, \emph{AnimeInbet}, is depicted in Figure \ref{fig:pipeline}.
Unlike existing frame interpolation methods that use raw raster images $I_0$ and $I_1$, we process vector graphs $G_0=\{V_0,T_0\}$ and $G_1=\{V_1,T_1\}$ instead.
The vertex coordinates in the images are represented by $V\in \mathbb R^{K\times2}$, and the binary adjacency matrix is denoted by $T\in {0,1}^{K\times K}$, where $K$ denotes the number of vertices. 
The goal is to generate the intermediate graph $G_t$ at time $t\in(0, 1)$.
To this end, we first design a CNN-based vertex geometric embedding to encode $V_0$ and $V_1$ to features $F_0$ and $F_1$
, respectively, as detailed in Section \ref{subsec:desc}.
Along with the embeddings, a vertex correspondence Transformer is proposed to aggregate the mutuality of vertex features to $\hat F_0$ and $\hat F_1$ by alternating self- and cross-attention layers (Section \ref{subsec:vct}).
The aggregated features are used to compute the correlation matrix $\mathcal C \in \mathbb R^{K_0\times K_1}$ and to induce the vertex matching by row-wise and column-wise argmax.
 In cases where vertices are occluded during large motion, we adopt a self-attention-based layer to propagate the vertex shifts from matched vertices to the unmatched, 
and obtain repositioning vectors $r_0\in \mathbb R^{K_0\times 2}$ and $r_1\in \mathbb R^{K_1\times 2}$ for all vertices (Section \ref{subsec:rp}).
Finally, we superpose the two input graphs based on the predicted correspondence, and we further refine the output by predicting visibility maps $m_0\in\{0,1\}^{K_0}$ and $m_1\in\{0,1\}^{K_1}$ to mask off those vertices of $V_0$ and $V_1$ that disappear in the intermediate frame, respectively, to obtain the final inbetweened line drawing $G_t$, as explained in Section \ref{subsec:fusion}.

\begin{figure}[t]

    \setlength{\tabcolsep}{1.5pt}
    \centering
    \includegraphics[width=0.96\linewidth]{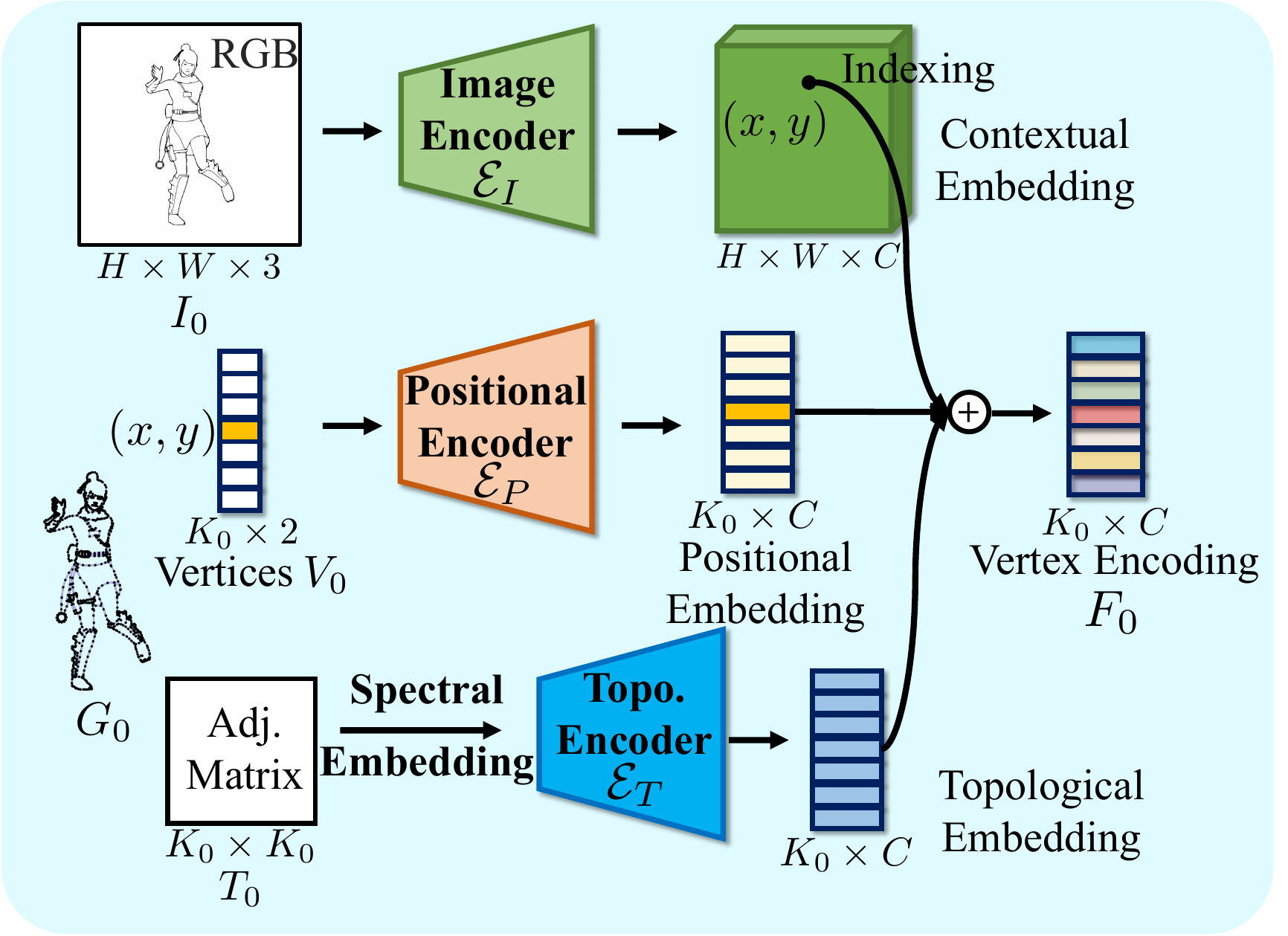}

    \caption{\small{\textbf{Vertex Geometric Embedding.} The goal is to obtain discriminative and meaningful features to describe each vertex.}} 
    \label{fig:descriptor}

\end{figure}


\noindent{\textbf{Geometrizing Line Drawings.}}
%
The process of creating artwork has become largely digital, allowing for direct export in vectorized format.
However, for line drawings that only appear in raster images, there are various commercial software and open-source research projects available \cite{zhang2009vectorizing,yao2016manga,virtualskethcer,liu2022end} that can be used to convert the raster images into the required vectorized input format. We will ablate the performance of line vectorization in our experiments.

\subsection{Vertex Geometric Embedding}
\label{subsec:desc}


Discriminative features for each vertex are desired to achieve accurate graph matching.
%
Line graphs are different from general graphs as the spatial position of endpoint vertices, in addition to the topology of connections, determines the geometric shape of the line.
%
The geometric graph embedding for line art is hence designed to comprise three parts: \textbf{1) image contextual embedding}, \textbf{2) positional embedding}, and \textbf{3) topological embedding}, as shown in Figure \ref{fig:descriptor}.

For image contextual embedding, we use a 2D CNN $\mathcal E_I$ to extract deep contextual features within the same size of the input raster image $I$.
Then, for each vertex $V_0[i] := (x,y)$ we index feature $\mathcal E_I(I)\left[(x,y)\right]$ as the image embedding for the $i$-th vertex.
As to the positional embedding, we employ a 1D CNN $\mathcal E_P$ to map each vertex coordinate $(x,y)$ to a $C$-dimensional feature.  
To include the topological information into a lower dimensional feature, we first conduct spectral embedding \cite{belkin2003laplacian} $\mathcal S$ on the binary adjacency matrix $T$, which involves an eigenvector decomposition on the  Laplacian matrix of the graph, then feed the spectral embedding to a subsequent 1D CNN $\mathcal E_T$.
The final geometric graph embedding is formulated as 
\begin{equation}
    F_0 = \mathcal E_I\left(I_0\right)\left[V_0\right] + \mathcal E_P\left(V_0\right) + \mathcal E_T\left(\mathcal S\left(T_0\right)\right).
\end{equation}
We obtain $F_1$ in the same way.

\begin{figure}[t]

    \setlength{\tabcolsep}{1.5pt}
    \centering
    \includegraphics[width=0.98\linewidth]{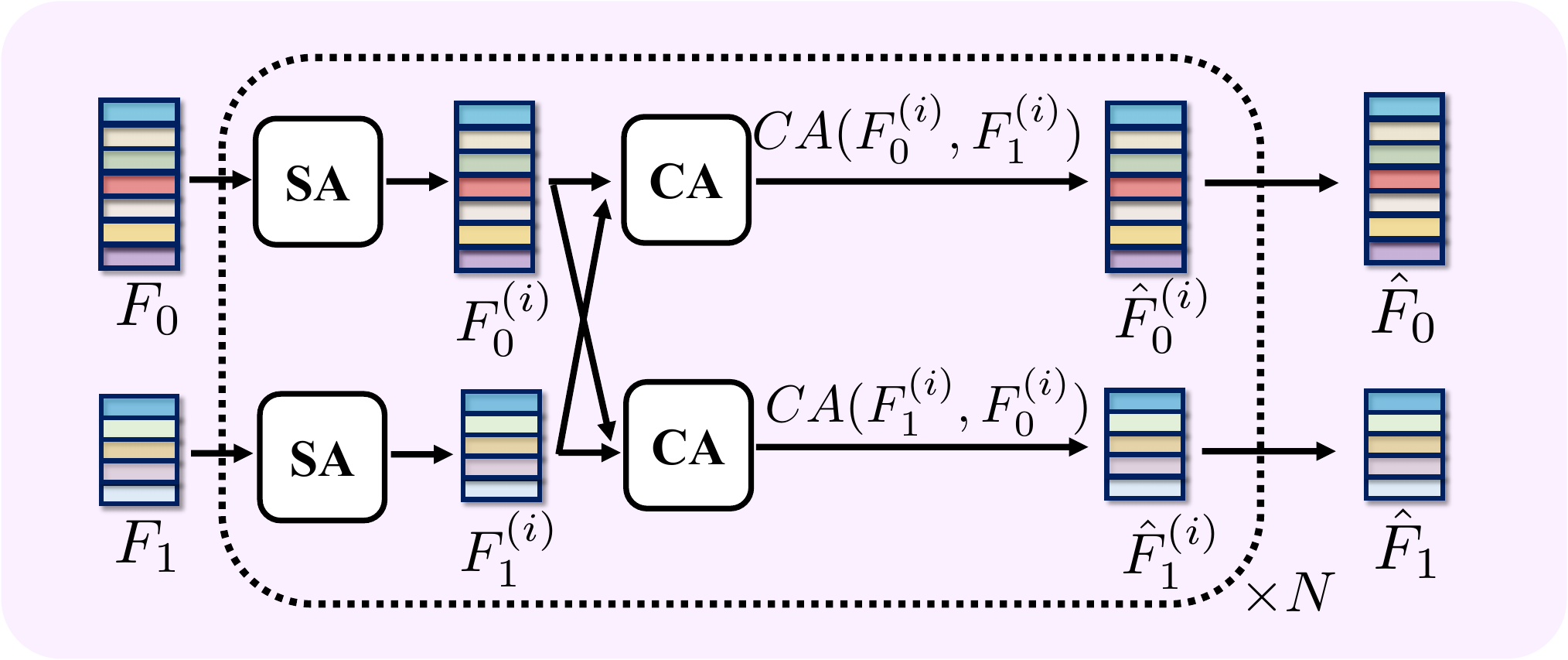}

    \vspace{-2pt}
    \caption{\small{\textbf{Vertex Correspondence Transformer.}} SA and CA represent self-attention and cross-attention, respectively.} 
    
    \label{fig:transformer}

\end{figure}

\subsection{Vertex Correspondence Transformer}
\label{subsec:vct}

We use geometric features $F_0$ and $F_1$ to establish a vertex-wise correspondence between $G_0$ and $G_1$.
Specifically, we compute a correlation matrix between vertex features and identify the matching pair as those with the highest value across both the row and the column of the matrix. 
Prior to this step, we apply a Transformer that aggregates the mutual consistency both intra- and inter-graph.

\noindent\textbf{Mutual Aggregation.}
Following \cite{sarlin2020superglue,sun2021loftr}, we employ a cascade of alternating self- and cross-attention layers to aggregate the vertex feature.
In a self-attention layer, all queries, keys and values are derived from the single source feature, 
\begin{equation}
    SA(F_0) = \text{softmax}\left( \frac{\mathcal Q (F_0) \mathcal K^T(F_0)}{\sqrt C}\right)\mathcal V(F_0),
\end{equation}
where $\mathcal Q$, $\mathcal K$ and $\mathcal V$ represent MLPs for query, key and value, respectively;
while in the cross-attention layer, the keys and values are computed from another feature: 
\begin{equation}
    CA(F_0, F_1) = \text{softmax}\left( \frac{\mathcal Q (F_0) \mathcal K^T(F_1)}{\sqrt C}\right)\mathcal V(F_1).
\end{equation}
After $N$ layers of rotating self- and cross-attention layers as shown in Figure \ref{fig:transformer}, we obtain aggregated feature $\hat F_0$ and $\hat F_1$. 
In the aggregation, each vertex is represented as an attentional pooling of all other vertices within the same graph and across the two graphs
achieving a full fusion of information with mutual dependencies.

\noindent\textbf{Correlation Matrix  and Vertex Matching.}
We compute the correlation matrix $\mathcal P$ as 
$\mathcal P = \frac{\hat F_0 \hat F_1^T}{\sqrt C}$.
We further apply a differentiable optimal transport ($OT$) \cite{sarlin2020superglue} to improve the dual selection consistency and obtain $\hat{\mathcal P} = OT(\mathcal P)$.
Then, we predict the one-way matching from $G_0$ to $G_1$ and vice versa as $\arg\max$ indices across rows and columns:
\begin{equation}
\label{eq:prediction}
\left\{ 
        \begin{array}{lll}
            \mathcal M_{0\to1}=\{(i,j) | j = \arg\max\hat{\mathcal P}_{i,:}, i=0, ..., K_0-1\} && \\
              \mathcal M_{1\to0}=\{(i,j) | i = \arg\max\hat{\mathcal P}_{:,j}, j=0, ..., K_1-1\}. &&  
        \end{array}
        \right.
\end{equation}
A vertex pair is selected into the final correspondence if it is mutually consistent and its correlation value is larger than $\theta$:
\begin{equation}
\label{eq:final_match}
    \hat{\mathcal M} = \left\{(i,j)| (i,j)\in\mathcal M_{0\to1} \cap M_{1\to0}, \hat{\mathcal P}_{i,j} > \theta\right\}.
\end{equation}
Otherwise, vertices will be considered to be occluded.

\begin{table*}
\renewcommand\arraystretch{0.86}
\caption{\small\textbf{Quantitative evaluations of state-of-the-art frame interpolation methods using Chamfer Distance} (reported in units of $\times 10^{-5}$, with lower values indicating better performance). The first place and runner-up are highlighted in bold and underlined, respectively.
}
\vspace{-4pt}
  
  \centering
  \small
  \setlength{\tabcolsep}{4mm}
{
  \begin{tabular}{l  c c c c c c c  c   }
    \toprule
    &  \multicolumn{4}{c}{  Validation Set } & \multicolumn{4}{c}{ Test Set }             \\

     \cmidrule(r){2-5}  \cmidrule(r){6-9} 
   Method & gap = 1  & gap = 5  & gap = 9 & Avg. & gap = 1  & gap = 5  & gap =  9  & Avg.\\
      \midrule
    VFIformer~\cite{lu2022video} & \ \  7.82& 26.04& 50.71 & 28.19 & \ \ 7.62 & 27.55 & 50.68 & 28.62 \\ 
    RIFE~\cite{huang2022rife} &\ \ 5.02 & 27.79 & 49.81 & 27.54 & \ \ 5.85 & 28.91 & 51.08 & 28.61\\
    EISAI~\cite{chen2022improving} &\ \ 5.66 & 27.64 & 49.43  & 27.57 & \ \ 6.02 & 29.14 & 52.36   & 29.17\\
    FILM~\cite{reda2022film} & \ \ 3.18 & 16.84 & 30.74 & 16.92 &\ \ 3.50&17.94&33.51 & 18.31 \\
    \midrule
    \textbf{\textit{AnimeInbet}} (ours) & \ \ \textbf{2.20} & \textbf{11.12} & \textbf{21.27}&  \textbf{11.53} & \ \ \textbf{2.80} & \textbf{12.69} & \textbf{23.21} & \textbf{12.90}\\
    \textbf{\textit{AnimeInbet-VS}}{(ours)}  & \ \ \underline{2.62} & \underline{11.43}  & \underline{22.36} &  \underline{12.14} & \ \ \underline{3.44} & \underline{13.41}  &  \underline{23.67} & \underline{13.51}\\
    \bottomrule
  \end{tabular}
  }
  \label{table:quantitative_eval}
\end{table*}

\subsection{Repositioning Propagation}
\label{subsec:rp}
%
Fused vertices $(i, j)$ from vertex correspondence can be linearly relocated to $tV_0[i]+(1-t)V_1[j]$ in intermediate graph $G_t$ based on time $t$.
%
However, the positions of the unmatched vertices in $G_t$ are still unknown. 
To reposition these vertices, we design an attention-based scheme similar to Xu~\etal~\cite{xu2022gmflow} to predict bidirectional shift vectors $r_{0\to1}$ and $r_{1\to0}$ for $V_0$ and $V_1$, respectively.
Formally,
\begin{equation}
\label{eq:prediction}
\left\{ 
        \begin{array}{lll}
            r_{0\to 1} = \text{softmax}\left(\frac{\hat F_0 \hat F_0^T}{\sqrt C}\right)\left(\text{softmax}(\hat{\mathcal P})V_1 - V_0\right) \\
            r_{1\to 0} = \text{softmax}\left(\frac{\hat F_1 \hat F_1^T}{\sqrt C}\right)\left(\text{softmax}(\hat{\mathcal P^T})V_0 - V_1\right).  
        \end{array}
        \right.
\end{equation}
We then compute the final repositioning vectors as follows:
\begin{equation}
\label{eq:prediction}
r_{0}[i] = \left\{ 
        \begin{array}{lll}
           V_1[j] - V_0[i], & \text{if } \,\,\, \exists \,\,j \,\,\, s.t.\,\, (i,j) \in \hat{\mathcal M}, \\
           r_{0\to1}[i], & \text{otherwise},
        \end{array}
        \right.
\end{equation}
while $r_1$ is computed in a similar way. 

In this step, the motion vector $r_{0\to1}$ of an unmatched vertex $V_0[i]$ is computed as a softmax average of shifts to all vertices in $G_1$, \ie, $\text{softmax}(\hat{\mathcal P}_{i,:})V_1 - V_0$. It is then refined by attention pooling from matched vertices, based on self-similarity given by $\hat F_0 \hat F_0^T/\sqrt C$.
Vertices are reasonably repositioned in the new vector graph after this step.

\subsection{Visibility Prediction and Graph Fusion}
\label{subsec:fusion}

%
To handle occlusions in the source line arts, we use a three-layer MLP to predict binary visibility maps $m_0$ and $m_1$ for the input graphs, obtained as $m_0=\text{MLP}(\hat F_0)$ and $m_1=\text{MLP}(\hat F_1)$.
Then, we merge the vertices to $V_t$ in the two graphs according to the following rule:

\begin{equation}
\begin{aligned}
    V_t&=\left\{(1-t)V_0[i]+tV_1[j] \, \Big| \, (i,j)\in \hat{\mathcal M}\right\} \\
    &\cup \left\{V_0[i] + t\cdot r_0[i] \, \Big| \, i \notin \hat{\mathcal M}, m_0[i] = 1  \right\}\\  
    &\cup \left\{V_1[j] + (1-t) r_1[j] \, \Big| \, j \notin \hat{\mathcal M}, m_1[j] = 1 \right\}, 
    \end{aligned}
\end{equation}
where we implement the repositioning that is compatible with arbitrary time $t\in(0,1)$.
As to $T_t$, we union all original connections if both endpoint vectors are both visible in $G_t$. Or formally, $T_t[\widetilde i][\widetilde j]=T_t[\widetilde j][\widetilde i]=1$ if $T_0[i][j]=1$ or $T_1[i][j]=1$, where $(i,j)$ and $(\widetilde i, \widetilde j)$ are the vertex indices in the original graph and the merged one.

\subsection{Learning}
\label{subsec:fusion}

The training objective of \textit{AnimeInbet} composes of three terms:
    $\mathcal L = \mathcal L_c + \mathcal L_r + \mathcal L_m$,
where the $\mathcal L_c$, $\mathcal L_r$ and $\mathcal L_m$ are used to supervise the learning of vertex matching $\hat{\mathcal M}$, repositioning vectors $r_0$ and $r_1$, and visibility masks $m_0$ and $m_1$, respectively.
$\mathcal L_c$ is to enlarge the correlation values of ground truth pairs and is defined as:
\begin{equation}
    \mathcal L_c = -\frac{1}{|\mathcal{M}^{GT}|}\sum_{(i,j)\in \mathcal{M}^{GT}}\log \hat{\mathcal P}_{i,j},
\end{equation}
where $\mathcal{M}^{GT}$ is the ground truth matching labels. 
For $\mathcal L_r$ and $\mathcal L_m$, we regress $r_{0\to1}$, $r_{1\to0}$, $m_0$, and $m_1$ as follows:
\begin{equation}
    \begin{aligned}
        \mathcal L_r &= \frac{1}{K_0}\|r_{0\to1}-r_{0\to1}^{GT}\|_1 + \frac{1}{K_1}\|r_{1\to0}-r_{1\to0}^{GT}\|_1 \\
    \mathcal L_m &= \text{BCE}^w\left(\sigma(m_0), m_0^{GT}\right) + \text{BCE}^w\left(\sigma(m_1), m_1^{GT}\right),
    \end{aligned}
\end{equation}
where $\sigma$ represents the sigmoid function, and BCE$^w$ is the binary cross-entropy loss with bias weight $w$.
However, since the shift vectors of occluded vertices cannot be obtained directly by subtracting the matched vertices, we conduct a frame-by-frame backtrack to generate pseudo labels to support the point-wise supervision of the repositioning vector and visibility maps.

\noindent\textbf{Pseudo Labels of Repositioning and Visibility.}
Assume $G^{(0)}$  and $G^{(Z)}$ are the $0$-th and the $Z$-th  frames in a training sequence, which are used for two input line sources.
Although there can exist many unmatched vertices in the two graphs when the gap $Z$ is large, the matching rate between adjacent frames (gap = 0) is relatively high according to Table \ref{table:statistics}. 
Based on this, we iteratively backtrack a shift vector $r^{(z)}$ from the $G^{(Z)}$ to $G^{(0)}$:
\begin{equation}
    r^{(z)}[i] = 
    \left\{ 
        \begin{array}{lll}V^{(z+1)}[j] - V_{(z)}[i] + r^{(z+1)},\,\, \text{if} \,\, i,j \, \text{is matched} \\
        \frac{1}{|\mathcal N_i|} \sum_{k \in {\mathcal N_i}} r^{(z)}[k], \,\, \text{otherwise}
                \end{array}
        \right.
\end{equation}
where $\mathcal N_i$ regards to the neighbors of the $i$-th vertex in $G^{(z)}$ and $r^{(Z)}$ is initialized to be $0$.
%
%
The termination $r^{(0)}$ of the backtrack is regarded as the pseudo repositioning label $r_{0\to1}^{GT}$.
As to the visibility labels, we first deuce $r_{0\to t}^{GT}$ as above and compute $m_0^{GT}$ as
\begin{equation}
    m_0^{GT}[i] = 
    \left\{ 
        \begin{array}{lll}
        1, \,\,\, \text{if} \,\,\, V_0[i] + r_{0\to t}^{GT} \in \widetilde I_t,\\
        0, \,\,\, \text{otherwise},
                \end{array}
        \right.
\end{equation}
where $\widetilde I_t$ is $I_t$ dilated by a $3\times3$ kernel.
$r_{1\to0}^{GT}$ and $m_1^{GT}$ are computed in reversed order.


\begin{figure*}[t]

    \setlength{\tabcolsep}{1.5pt}
    \centering
    \includegraphics[width=0.996\linewidth]{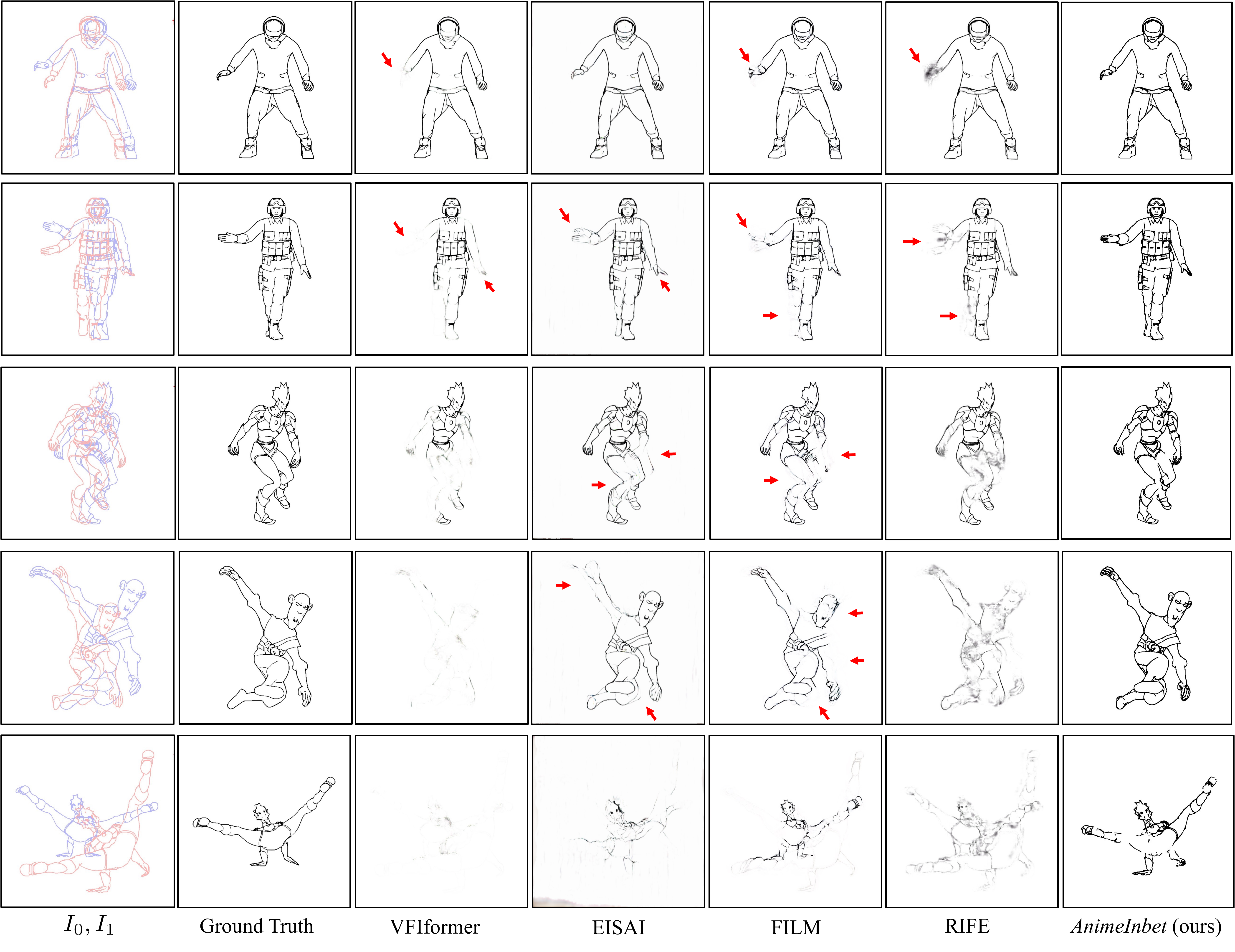}

    \vspace{-6pt}
    \caption{\small{\textbf{Inbetweening results on \textit{MixamoLine240} test set.} Examples are arranged from small (top) to large (bottom) motion magnitudes.
    %
    }} 
    \label{fig:comparison}

\end{figure*}

\section{Experiments}
\label{sec:experiment}
\noindent\textbf{Implementation Details.}
In the vertex geometric embedding module, the image encoder $\mathcal E_I$ is implemented as a three-layer 2D CNN, while the positional encoder $\mathcal E_P$ and the topological encoder $\mathcal E_T$ are 1D CNNs with a kernel size of $1$.
Encoding feature $C$ is $128$ in our experiments.
Before feeding vertex coordinates $V$ into $\mathcal E_P$, $V$ are first normalized to the scale between $(-1, 1)$; the dimension of the spectral embedding feature is $64$.
Threshold $\theta$ in Equation \ref{eq:final_match} is $0.2$.
In both training and evaluation, intermediate time $t$ is $0.5$, which regards the center frame between $I_0$ and $I_1$.
The detailed network structures are provided in the supplementary file.
We use Adam \cite{kingma2014adam} optimizer with a learning rate of $1\times  10^{-4}$ to train the \textit{AnimeInbet} for $70$ epochs, where we first solely supervise the network using the correspondence loss $\mathcal L_c$ for the $50$ epochs, and  then adopt the full loss $\mathcal L$ for the rest $20$ epochs.
Bias weight $w$ in $\mathcal L_m$ is $0.2$.
Since vertex numbers differ in frames, we feed one pair of input frames each time but adopt gradient accumulation for a mini-batch size of $8$.
The model is trained with an NVIDIA Tesla V100 GPU for about five days.
During the test, $G_t$ is visualized as a raster image by \texttt{cv2.line} function with a line width of 2 pixels. 
%
We evaluate our model on both ground truth vectorization labels   (noted as ``\textit{AnimeInbet}'') and those vectorized from VirtualSketcher \cite{virtualskethcer} (noted as ``\textit{AnimeInbet-VS}'', to simulate the cases when input anime drawing are vector and raster, respectively.


\noindent\textbf{Evaluation Metric.}
Following~\cite{narita2019optical,chen2022improving}, we adopt the chamfer distance (CD) as the evaluation metric, which has been initially introduced to measure the similarity between two point clouds.
Formally, CD is computed as:
\begin{equation}
    \begin{aligned}
        CD(I_{t},I_{t}^{GT}) = \frac{1}{HWd}\sum (I_{t}\textit{DT}(I_{t}^{GT}) + I_{t}^{GT}\textit{DT}(I_{t})),
    \end{aligned}
\end{equation}
where $I_t$ and $I_t^{GT}$ are predicted binary lines and ground truth,  while $H$, $W$ and $d$ are image height, width, and a search diameter \cite{chen2022improving}, respectively. \textit{DT} denotes the Euclidean distance transform. To transfer predicted raster images into binary sketches, we threshold pixels smaller than 0.99 times the maximum value to 0.

\subsection{Comparison to Existing Methods}
%
Since there is no existing geometrized line inbetweening study that we can directly compare  our proposed model with, we set several state-of-the-art raster-image-based  frame interpolation methods as baselines, including VFIformer \cite{lu2022video}, RIFE \cite{huang2022rife}, EISAI \cite{chen2022improving}, FILM \cite{reda2022film}.
Specifically, EISAI is originally intended for 2D animation and embeds an optical flow-based contour aggregator.
We test each model's performance on frame pairs within frame gaps of 1, 5 and 9, respectively.
For fairness, we finetune each compared method on the training set of \textit{MixiamoLine240} with relative frame gaps using a learning rate of $1\times 10^{-6}$ for five epochs.

As shown in Table \ref{table:quantitative_eval}, our \textit{AnimeInbet} favorably outperforms all compared methods on both the validation set and the test set of \textit{MixamoLine240}.
On the validation set, our approach achieves an average CD value of $11.53$, representing a significant improvement over the best-performing compared method, FILM, with over $30\%$ enhancement. 
Upon closer inspection, the advantage of \textit{AnimeInbet} becomes more pronounced as the frame gap increases ($0.98$, $5.72$ and $9.47$ for gaps of 1, 5, and 9, respectively), indicating that our method is more robust in handling larger motions.
On the test set, our method maintains its lead over the other compared methods, with improvements of $0.70$ ($20\%$), $5.25$ ($29\%$), and $10.30$ ($31\%$) from the best-performing compared method FILM for the frame gaps of 1, 5, and 9, respectively.
Given that both the characters and actions in the test set are new, our method's superiority on the test set provides more convincing evidence of its advantages over the existing frame interpolation methods.

\begin{figure}[t]

    \setlength{\tabcolsep}{1.5pt}
    \centering
    \includegraphics[width=0.94\linewidth, trim=0pt 0pt 0pt 0pt, clip]{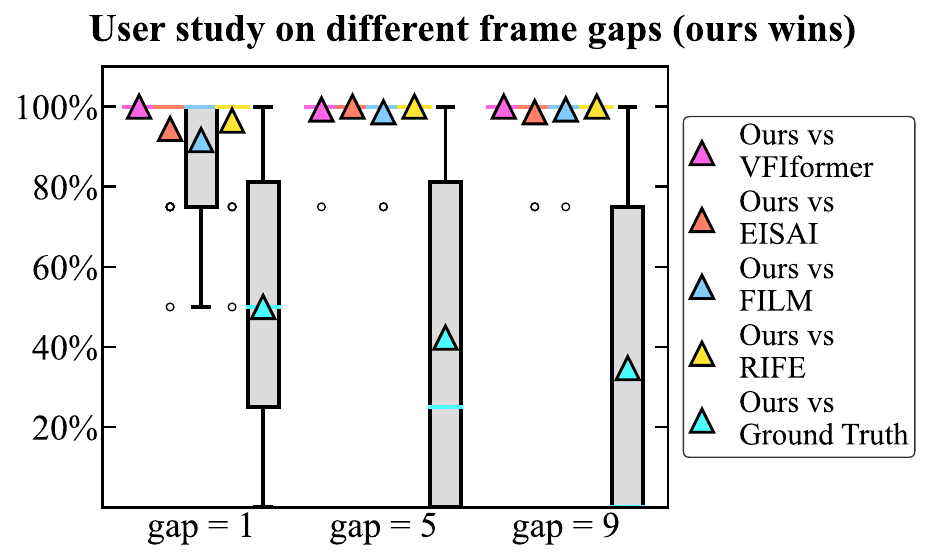}

    \vspace{-2pt}
    \caption{\small{\textbf{Statistics of user study.} In the boxplot, triangles and colored lines represent mean and median values, respectively. Circles are outliers beyond $1.5\times$ interquartile range ($3\sigma$ in a normal distribution).}} 
    \label{fig:user}

\end{figure}

\begin{figure*}[t]

    \setlength{\tabcolsep}{1.5pt}
    \centering
    \includegraphics[width=0.94\linewidth, trim=0pt 0pt 0pt 5pt, clip]{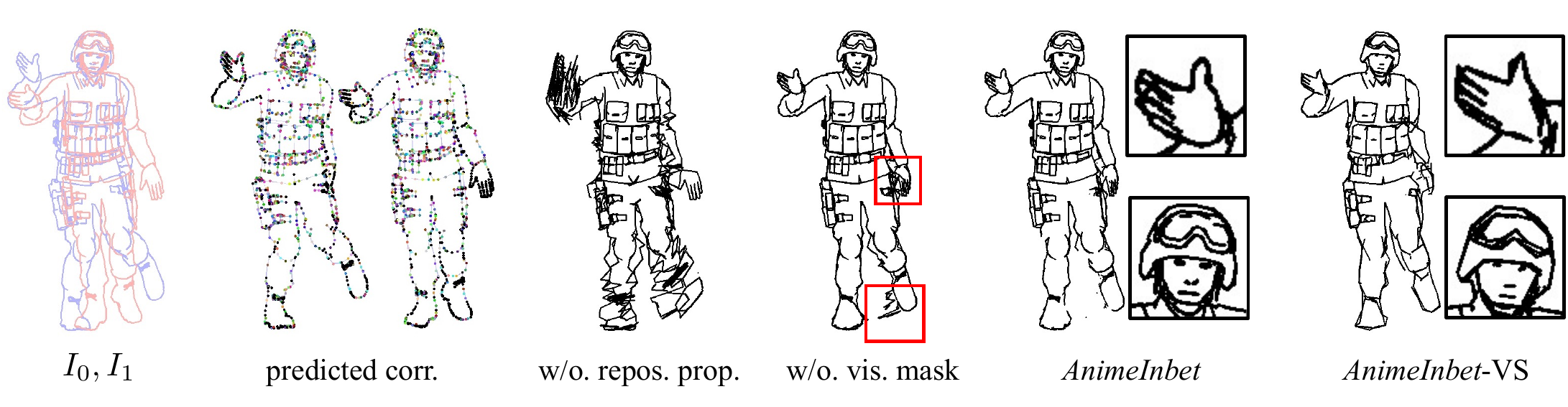}

    \vspace{-6pt}
    \caption{\small{\textbf{Visualization of ablation study.} In predicted correspondence, matched vertices are marked in the same colors, while unmatched are black (please zoom in).}} 
    \label{fig:ablation_user}

\end{figure*}


To illustrate the advantages of our method, we present several inbetweening results in Figure \ref{fig:comparison}. We arranged these examples in increasing levels of difficulty from top to bottom.
When the motion is simple, compared methods can interpolate a relatively complete shape of the main body of the drawing. However, they tend to produce strong blurring (RIFE) or disappearance (VFIformer, EISAI, and FILM) of noticeable moving compositions (indicated by red arrows). In contrast, our method maintains a concise line structure in these key areas.
When the input frames involve the whole body's movement within large magnitudes, the intermediate frames predicted by the compared methods become indistinguishable and patchy, rendering the results invalid for further use. However, our \textit{AnimeInbet} method can still preserve the general shape in the correct positions, even with a partial loss of details, which can be easily fixed with minor manual effort.


\noindent{\textbf{User Study.}} To further evaluate the visual performance of
our methods, we conduct a user study among 36 participants.
For each participant, we randomly show  60 pairs, each composed of a result of \textit{AnimeInbet} and that of a compared method, and ask the participant to select the better.
To allow participants to take  temporal consistency into the decision, we display these results in GIF formats formed by triplets of input frames and the inbetweened one.
The winning rates of our method  
are shown in Figure \ref{fig:user}, where \textit{AnimeInbet} wins over $92\%$ versus the compared methods.
Notably, for ``gap = 5'' and  ``gap = 9'' slots, the winning rates of our methods are close to $100\%$ with smaller  deviations than ``gap = 1'', suggesting the advantages of our method on cases within large motions.  

\begin{table}
\caption{\small\textbf{Ablation study on vertex encoding.}
}
\vspace{-4pt}
  
  \centering
  \small
{
  \begin{tabular}{c c c  c c c  }
    \toprule
    
      $\mathcal E_I$ & $\mathcal E_P$ & $\mathcal E_T$ & Acc. (\%) & Valid Acc. (\%)  & CD ($\downarrow$)\\
      \midrule
      \cmark & \xmarkg  & \xmarkg & 51.66 & 31.01 & 12.30\\
      \cmark & \cmark & \xmarkg & 61.87 & 55.62 & 11.55\\
      \cmark &\xmarkg & \cmark  & 59.28 & 45.45  & 11.86 \\
    \cmark & \cmark & \cmark & \textbf{65.51} & \textbf{61.28} & \textbf{11.12} \\ 
    \bottomrule
  \end{tabular}
  }
  \label{table:ablation_feature}
\end{table}

\begin{table}
\caption{\small\textbf{Ablation study on repositioning and visibility mask.}
}
\vspace{-4pt}
  
  \centering
  \small
{
  \begin{tabular}{l    c  }
    \toprule
    Method & CD ($\downarrow$) \\
    \midrule
    w/o. repositioning propagation & 23.62\\
    w/o. visibility mask & 12.81  \\ 
    full model & 11.12 \\ 
     
    \bottomrule
  \end{tabular}
  }
  \label{table:ablation_2}
\end{table}

\subsection{Ablation Study}
\noindent{\textbf{Embedding Features.}}
To investigate the effectiveness of the three types of embeddings mentioned in Section \ref{subsec:desc}, we trained several variants by removing the corresponding modules.
As shown in Table \ref{table:ablation_feature}, for each variant, we list the matching accuracy for all vertices (``Acc.''), the accuracy for non-occluded vertices (``Valid Acc.'') and the final CD values of inbetweening on the validation set (gap = 5).
If removing the positional embedding $\mathcal E_P$, the ``Valid Acc.'' and the CD value drop $15.83\%$ and $0.74$, respectively; while the lacking of topological embedding $\mathcal E_T$ lowers ``Valid Acc.'' by $5.66\%$ and worsens CD by $0.43$, which reveals the importance of these two components.

\noindent{\textbf{Repositioning Propagation and Visibility Mask.}}
We demonstrate the contribution of repositioning propagation (prepos. prop.)  and visibility mask (vis. mask) both quantitatively  and qualitatively.
As shown in Table \ref{table:ablation_2}, without repositioning propagation, the CD value will be sharply worsened by $12.50$ ($112\%$), while the lacking of visibility mask will also make a drop of $1.69$ ($15\%$). 
An example is shown in Figure \ref{fig:ablation_user}, where ``w/o. repos. prop.'' appears within many messy lines due to undefined positions for those unmatched vertices, while ``w/o. vis. mask'' shows some redundant segments (red box) after repositioning; 
the complete \textit{AnimeInbet} can resolve these issues and produce a clean yet complete result.

\noindent{\textbf{Geometrizor.}} 
%
As shown in Table \ref{table:quantitative_eval}, the quantitative metrics of \textit{AnimeInbet-VS} are generally worse by around $0.6$ compared to \textit{AnimeInbet}.
This is because VirtualSketcher \cite{virtualskethcer} does not vectorize the line arts as precisely as the ground truth labels (average vertex number 587 \textit{vs} 1,351).
As shown in Figure \ref{fig:ablation_user}, the curves in ``\textit{AnimeInbet-VS}'' become sharper and lose some details, which decreases the quality of the inbetweened frame. Using a more accurate geometrizer would lead to higher quality inbetweening results for raster image inputs.


\noindent{\textbf{Data Influence.}
As mentioned in Section \ref{sec:data}, we created a validation set composed of 20 sequences of unseen characters but seen actions, 20 of unseen actions but seen characters and 4 of unseen both to explore the influence on data.
Our experiment finds that whether the characters or the actions are seen does not fundamentally influence the inbetweening quality, while the motion magnitude is the key factor.
%
As shown in Table \ref{table:ablation_data}, the CD value of unseen characters is $14.70$, which is over $47\%$ worse than that of unseen both due to larger vertex shifts   ($44.59$ \textit{vs} $29.62$), while the difference between the CD values of unseen actions and unseen both is around 10\% under similar occlusion rates and shifts.

\begin{table}
\caption{\small\textbf{Ablation study on data influence.} 
}
\vspace{-4pt}
  
  \centering
  \small
{
  \begin{tabular}{l c c   c  }
    \toprule
    Validation data (gap = 5) & Occ. (\%) & Shift & CD ($\downarrow$) \\
    \midrule
    Unseen characters ($2\times10$) & 34.30 & 44.59 & 14.70\\
    Unseen actions ($10\times2$) & 37.71 & 31.53 & \ \ 8.98 \\
    Unseen both ($2\times2$) & 34.10 & 29.62 & \ \ 9.98\\
    
    \bottomrule
  \end{tabular}
  }
  \label{table:ablation_data}
\end{table}


\section{Conclusion}
\label{sec:conclusion}

In this study, we address the practical problem of cartoon line inbetweening and propose a novel approach that treats line arts as geometrized vector graphs. Unlike previous frame interpolation tasks on raster images, our approach formulates the inbetweening task as a graph fusion problem with vertex repositioning. 
We present a deep learning-based framework called \textit{AnimeInbet}, which shows significant gains over existing methods in terms of both quantitative and qualitative evaluation. 
To facilitate training and evaluation on cartoon line inbetweening, we also provide a large-scale geometrized line art dataset, \textit{MixamoLine240}. Our proposed framework and dataset facilitate a wide range of applications, such as anime production and multimedia design, and have significant practical implications.

\noindent{\textbf{Acknowledgement.} 
This research is supported by the National Research Foundation, Singapore under its AI
Singapore Programme (AISG Award No: AISG-PhD/2021-01-031[T]).  It is also supported under the RIE2020 Industry Alignment Fund Industry Collaboration Projects (IAF-ICP) Funding Initiative, as well as cash and in-kind contribution from the industry partner(s). 
This study is partially supported by NTU NAP, MOE AcRF Tier 1 (2021-T1-001-088).}

{\small
\bibliographystyle{ieee_fullname}
\bibliography{egbib.bib}
}

\end{document}